\pdfoutput=1 
\documentclass[11pt,a4paper]{article}
\usepackage[hyperref]{acl2020}
\usepackage{times}
\usepackage{latexsym}

\usepackage{microtype}

\aclfinalcopy 

\usepackage[utf8]{inputenc}
\usepackage{multirow}
\usepackage{booktabs}

\title{Do all Roads Lead to Rome?\\Understanding the Role of Initialization in Iterative Back-Translation}

\author{Mikel Artetxe$^{1}$, \ Gorka Labaka$^{1}$, \ Noe Casas$^{2}$, \ Eneko Agirre$^{1}$ \\
$^1$IXA Group, University of the Basque Country (UPV/EHU) \\
$^2$TALP Research Center, Universitat Politècnica de Catalunya \\
\texttt{\{mikel.artetxe,gorka.labaka,e.agirre\}@ehu.eus} \\ 
\texttt{noe.casas@upc.edu}
}

\date{}

\begin{document}
\maketitle
\begin{abstract}
Back-translation provides a simple yet effective approach to exploit monolingual corpora in Neural Machine Translation (NMT). Its iterative variant, where two opposite NMT models are jointly trained by alternately using a synthetic parallel corpus generated by the reverse model, plays a central role in unsupervised machine translation. In order to start producing sound translations and provide a meaningful training signal to each other, existing approaches rely on either a separate machine translation system to warm up the iterative procedure, or some form of pre-training to initialize the weights of the model. In this paper, we analyze the role that such initialization plays in iterative back-translation. Is the behavior of the final system heavily dependent on it? Or does iterative back-translation converge to a similar solution given any reasonable initialization? Through a series of empirical experiments over a diverse set of warmup systems, we show that, although the quality of the initial system does affect final performance, its effect is relatively small, as iterative back-translation has a strong tendency to convergence to a similar solution. As such, the margin of improvement left for the initialization method is narrow, suggesting that future research should focus more on improving the iterative mechanism itself.
\end{abstract}

\section{Introduction}
\label{sec:introduction}

Back-translation \citep{sennrich2016improving} allows to naturally exploit monolingual corpora in Neural Machine Translation (NMT) by using a reverse model to generate a synthetic parallel corpus. Despite its simplicity, this technique has become a key component in state-of-the-art NMT systems. For instance, the majority of WMT19 submissions, including the best performing systems, made extensive use of it \citep{barrault2019findings}.

While the synthetic parallel corpus generated through back-translation is typically combined with real parallel corpora, iterative or online variants of this technique also play a central role in unsupervised machine translation \citep{artetxe2018usmt,artetxe2018unmt,artetxe2019effective,lample2018unsupervised,lample2018phrase,marie2018unsupervised,conneau2019crosslingual,song2019mass,liu2020multilingual}. In iterative back-translation, both NMT models are jointly trained using synthetic parallel data generated on-the-fly with the reverse model, alternating between both translation directions iteratively. While this enables fully unsupervised training without parallel corpora, some initialization mechanism is still required so the models can start producing sound translations and provide a meaningful training signal to each other. For that purpose, state-of-the-art approaches rely on either a separately trained unsupervised Statistical Machine Translation (SMT) system, which is used for warmup during the initial back-translation iterations \citep{marie2018unsupervised,artetxe2019effective}, or large-scale pre-training through masked denoising, which is used to initialize the weights of the underlying encoder-decoder \citep{conneau2019crosslingual,song2019mass,liu2020multilingual}.

In this paper, we aim to understand the role that the initialization mechanism plays in iterative back-translation. For that purpose, we mimic the experimental settings of \citet{artetxe2019effective}, and measure the effect of using different initial systems for warmup: the unsupervised SMT system proposed by \citet{artetxe2019effective} themselves, supervised NMT and SMT systems trained on both small and large parallel corpora, and a commercial Rule-Based Machine Translation (RBMT) system. Despite the fundamentally different nature of these systems, our analysis reveals that iterative back-translation has a strong tendency to converge to a similar solution. Given the relatively small impact of the initial system, we conclude that future research on unsupervised machine translation should focus more on improving the iterative back-translation mechanism itself.

\section{Iterative back-translation}
\label{sec:backtranslation}

We next describe the iterative back-translation implementation used in our experiments, which was proposed by \citet{artetxe2019effective}. Note, however, that the underlying principles of iterative back-translation are very general, so our conclusions should be valid beyond this particular implementation.

The method in question trains two NMT systems in opposite directions following an iterative process where, at every iteration, each model is updated by performing a single pass over a set of $N$ synthetic parallel sentences generated through back-translation. After iteration $a$, the synthetic parallel corpus is entirely generated by the reverse NMT model. However, so as to ensure that the NMT models produce sound translations and provide meaningful training signal to each other, the first $a$ warmup iterations progressively transition from a separate \textbf{initial system} to the reverse NMT model itself. More concretely, iteration $t$ uses $N_{init} = N \cdot \max (0, 1 - t / a)$ back-translated sentences from the reverse initial system, and the remaining $N - N_{initial}$ sentences are generated by the reverse NMT model. In the latter case, half of the translations use random sampling \citep{edunov2018understanding}, which produces more varied translations, whereas the other half are generated through greedy decoding, which produces more fluent and predictable translations. Following \citet{artetxe2019effective}, we set $N=1,000,000$ and $a=30$, and perform a total of 60 such iterations. Both NMT models use the big transformer implementation from Fairseq\footnote{\url{https://github.com/pytorch/fairseq}}, training with a total batch size of 20,000 tokens with the exact same hyperparameters as \citet{ott2018scaling}. At test time, we use beam search decoding with a beam size of 5.

\section{Experimental settings}
\label{sec:settings}

So as to better understand the role of initialization in iterative back-translation, we train different English-German models using the following \textbf{initial systems} for warmup:
\begin{itemize}
\item \textbf{RBMT}: We use the commercial Lucy LT translator \citep{alonso2003comprendium}, a traditional transfer-based RBMT system combining human crafted computational grammars and monolingual and bilingual lexicons.
\item Supervised \textbf{NMT}: We use the Fairseq implementation of the big transformer model using the same hyperparameters as \citet{ott2018scaling}. We train two separate models: one using the concatenation of all parallel corpora from WMT 2014, and another one using a random subset of 100,000 sentences. In both cases, we use early stopping according to the cross-entropy in newstest2013.
\item Supervised \textbf{SMT}: We use the Moses \citep{koehn2007moses} implementation of phrase-based SMT \citep{koehn2003statistical} with default hyperparameters, using FastAlign \citep{dyer2013simple} for word alignment. We train two separate models using the same parallel corpus splits as for NMT. In both cases, we use a 5-gram language model trained with KenLM \citep{heafield2013scalable} on News Crawl 2007-2013, and apply MERT tuning \citep{och2003MERT} over newstest2013.
\item \textbf{Unsupervised}: We use the unsupervised SMT system proposed by \citet{artetxe2019effective}, which induces an initial phrase-table using cross-lingual word embedding mappings, combines it with an n-gram language model, and further improves the resulting model through unsupervised tuning and joint refinement.
\end{itemize}

For each initial system, we train a separate NMT model through iterative back-translation as described in Section \ref{sec:backtranslation}. For that purpose, we use the News Crawl 2007-2013 monolingual corpus as distributed in the WMT 2014 shared task.\footnote{Note that the final systems do not see any parallel data during training, even if some initial systems are trained on parallel data. Thanks to this, we can measure the impact of the initial system in a controlled environment, which is the goal of the paper. In practical settings, however, better results could likely be obtained by combining real and synthetic parallel corpora.} Preprocessing is done using standard Moses tools, and involves punctuation normalization, tokenization with aggressive hyphen splitting, and truecasing.

We \textbf{evaluate} in newstest2014 using tokenized BLEU, and compare the performance of the different final systems after iterative back-translation and the initial systems used in their warmup.\footnote{Note that all systems use the exact same tokenization, so the reported BLEU scores are comparable among them.} However, this only provides a measure of the \textbf{quality} of the different systems, but not the similarity of the translations they produce. So as to quantify how similar the translations of two systems are, we compute their corresponding BLEU scores taking one of them as the reference. This way, we report the average \textbf{similarity} of each final system with the rest of final systems, and analogously for the initial ones. Finally, we also compute the similarity between each initial system and its corresponding final system, which measures how much the final solution found by iterative back-translation differs from the initial one.

\section{Results}
\label{sec:results}

Table \ref{tab:test} reports the test scores of different initial systems along with their corresponding final systems after iterative-backtranslation. As it can be seen, the standard deviation across final systems is substantially lower than across initial systems (1.7 vs 5.6 in German-to-English and 1.4 vs 4.9 in English-to-German), which shows that iterative back-translation tends to converge to solutions of a similar quality. This way, while the initial system does have certain influence in final performance, differences greatly diminish after applying iterative back-translation. For instance, the full NMT system is 13.4 points better than the RBMT system in German-to-English, but this difference goes down to 2.3 points after iterative back-translation.

\begin{table}[t]
\begin{center}
\begin{small}
  \addtolength{\tabcolsep}{-1.8pt}
  \begin{tabular}{lccccc}
    \toprule
    & \multicolumn{2}{c}{DE-EN} && \multicolumn{2}{c}{EN-DE} \\
    \cmidrule{2-3} \cmidrule{5-6}
    & init & final && init & final \\
    \midrule
    RBMT & 19.1 & 27.3 && 15.6 & 22.8 \\
    NMT (full) & 32.5 & 29.6 && 27.6 & 24.9 \\
    NMT (100k) & 15.2 & 25.0 && 12.5 & 20.8 \\
    SMT (full) & 25.5 & 28.3 && 20.5 & 23.3 \\
    SMT (100k) & 19.6 & 25.0 && 16.3 & 21.0 \\
    Unsupervised & 20.1 & 26.1 && 15.8 & 21.9 \\
    \midrule
    Average & 22.0 & 26.9 && 18.1 & 22.4 \\
    Standard dev. & 5.6 & 1.7 && 4.9 & 1.4 \\
    \bottomrule
  \end{tabular}
\end{small}
\end{center}
\caption{Test results using different initial systems for warmup (BLEU), before (init column) and after iterative back-translation (final column).}
\label{tab:test}
\end{table}

\begin{table}[t]
\begin{center}
\begin{small}
  \addtolength{\tabcolsep}{-1.8pt}
  \begin{tabular}{lccccccc}
    \toprule
    & \multicolumn{3}{c}{DE-EN} & & \multicolumn{3}{c}{EN-DE} \\
    \cmidrule{2-4} \cmidrule{6-8}
    & init & init & final && init & init & final \\
    & init & final & final && init & final & final \\
    \midrule
    RBMT & 23.8 & 27.6 & 48.0 && 20.0 & 25.0 & 41.7 \\
    NMT (full) & 28.5 & 42.2 & 50.4 && 25.0 & 36.3 & 43.8 \\
    NMT (100k) & 21.8 & 21.9 & 47.2 && 18.1 & 17.9 & 41.1 \\
    SMT (full) & 35.4 & 33.2 & 53.4 && 30.5 & 25.2 & 46.7 \\
    SMT (100k) & 31.5 & 27.4 & 48.4 && 28.0 & 22.4 & 42.4 \\
    Unsupervised & 28.4 & 31.7 & 48.3 && 24.3 & 24.2 & 41.8 \\
    \midrule
    Average & 28.2 & 30.7 & 49.3 && 24.3 & 25.1 & 42.9 \\
    \bottomrule
  \end{tabular}
\end{small}
\end{center}
\caption{Average similarity across initial systems and final systems, as well as each initial system and its corresponding final system (BLEU).}
\label{tab:similarity}
\end{table}

Interestingly, better initial systems do not always lead to better final systems. For instance, the initial RBMT system is weaker than both the unsupervised system and the small SMT system, yet it leads to a better final system after iterative back-translation. Similarly, the small SMT model is substantially better than the small NMT model in German-to-English (19.6 vs 15.2), yet they both lead to the exact same BLEU score of 25.0 after iterative back-translation. We hypothesize that certain properties of the initial system are more relevant than others and, in particular, our results suggest that the adequacy and lexical coverage of the initial systems has a larger impact than its fluency.

At the same time, it is remarkable that iterative back-translation has a generally positive impact, bringing an average improvement of 4.9 BLEU points for German-to-English and 4.3 BLEU points for English-to-German. Nevertheless, the full NMT system is a notable exception, as the final system learned through iterative back-translation is weaker than the initial system used for warmup. This reinforces the idea that iterative back-translation converges to a solution of a similar quality regardless of that of the initial system, to the extent that it can even deteriorate performance when the initial system is very strong.

So as to get a more complete picture of this behavior, Table \ref{tab:similarity} reports the average similarity between each final system and the rest of the final systems, and analogously for the initial ones. As it can be seen, final systems trained through iterative back-translation tend to produce substantially more similar translations than the initial systems used in their warmup (49.3 vs 28.2 for German-to-English and 42.9 vs 24.3 for English-to-German). This suggests that iterative back-translation does not only converge to solutions of similar quality, but also to solutions that have a similar behavior. Interestingly, this also applies to systems that follow a fundamentally different paradigm as it is the case of RBMT. In relation to that, note that the similarity of each final system and its corresponding initial system is rather low, which reinforces the idea that the solution found by iterative back-translation is not heavily dependent on the initial system.

\section{Related work}
\label{sec:related}

Originally proposed by \citet{sennrich2016improving}, back-translation has been widely adopted by the machine translation community \citep{barrault2019findings}, yet its behavior is still not fully understood. Several authors have studied the optimal balance between real and synthetic parallel data, concluding that using too much synthetic data can be harmful \citep{poncelas2018investigating,fadaee2018backtranslation,edunov2018understanding}. In addition to that, \citet{fadaee2018backtranslation} observe that back-translation is most helpful for tokens with a high prediction loss, and use this insight to design a better selection method for monolingual data. At the same time, \citet{edunov2018understanding} show that random sampling provides a stronger training signal than beam search or greedy decoding. Closer to our work, the impact of the system used for back-translation has also been explored by some authors \citep{sennrich2016improving,burlot2018using}, although the iterative back-translation variant, which allows to jointly train both systems so they can help each other, was not considered, and synthetic data was always combined with real parallel data.

While all the previous authors use a fixed system to generate synthetic parallel corpora,  \citet{hoang2018iterative} propose performing a second iteration of back-translation. Iterative back-translation was also explored by \citet{marie2018unsupervised} and \citet{artetxe2019effective} in the context of unsupervised machine translation, relying on an unsupervised SMT system \citep{lample2018phrase,artetxe2018usmt} for warmup. Early work in unsupervised NMT also incorporated the idea of on-the-fly back-translation, which was combined with denoising autoencoding and a shared encoder initialized through unsupervised cross-lingual embeddings \citep{artetxe2018unmt,lample2018unsupervised}. More recently, several authors have performed large-scale unsupervised pre-training through masked denoising to initialize the full model, which is then trained through iterative back-translation \citep{conneau2019crosslingual,song2019mass,liu2020multilingual}. Finally, iterative back-translation is also connected to the reconstruction loss in dual learning \citep{he2016dual}, which incorporates an additional language modeling loss and also requires a warm start.

\section{Conclusions}
\label{sec:conclusions}

In this paper, we empirically analyze the role that initialization plays in iterative back-translation. For that purpose, we try a diverse set of initial systems for warmup, and analyze the behavior of the resulting systems in relation to them. Our results show that differences in the initial systems heavily diminish after applying iterative back-translation. At the same time, we observe that iterative back-translation has a hard ceiling, to the point that it can even deteriorate performance when the initial system is very strong. As such, we conclude that the margin for improvement left for the initialization is rather narrow, encouraging future research to focus more on improving the iterative back-translation mechanism itself.

In the future, we would like to better characterize the specific factors of the initial systems that are most relevant. At the same time, we would like to design a simpler unsupervised system for warmup that is sufficient for iterative back-translation to converge to a good solution. Finally, we would like to incorporate pre-training methods like masked denoising into our analysis.

\section*{Acknowledgments}

This research was partially funded by a Facebook Fellowship, the Basque Government excellence research group (IT1343-19), the Spanish MINECO (UnsupMT TIN2017‐91692‐EXP MCIU/AEI/FEDER, UE), Project BigKnowledge (Ayudas Fundación BBVA a equipos de investigación científica 2018), the NVIDIA GPU grant program, Lucy Software / United Language Group (ULG), and the Catalan Agency for Management of University and Research Grants (AGAUR) through an Industrial Ph.D. Grant.

\bibliography{acl2020}

\begin{thebibliography}{24}
\expandafter\ifx\csname natexlab\endcsname\relax\def\natexlab#1{#1}\fi

\bibitem[{Alonso and Thurmair(2003)}]{alonso2003comprendium}
Juan~A Alonso and Gregor Thurmair. 2003.
\newblock The comprendium translator system.
\newblock In \emph{Proceedings of the Ninth Machine Translation Summit}.

\bibitem[{Artetxe et~al.(2018{\natexlab{a}})Artetxe, Labaka, and
  Agirre}]{artetxe2018usmt}
Mikel Artetxe, Gorka Labaka, and Eneko Agirre. 2018{\natexlab{a}}.
\newblock \href {https://doi.org/10.18653/v1/D18-1399} {Unsupervised
  statistical machine translation}.
\newblock In \emph{Proceedings of the 2018 Conference on Empirical Methods in
  Natural Language Processing}, pages 3632--3642, Brussels, Belgium.
  Association for Computational Linguistics.

\bibitem[{Artetxe et~al.(2019)Artetxe, Labaka, and
  Agirre}]{artetxe2019effective}
Mikel Artetxe, Gorka Labaka, and Eneko Agirre. 2019.
\newblock \href {https://doi.org/10.18653/v1/P19-1019} {An effective approach
  to unsupervised machine translation}.
\newblock In \emph{Proceedings of the 57th Annual Meeting of the Association
  for Computational Linguistics}, pages 194--203, Florence, Italy. Association
  for Computational Linguistics.

\bibitem[{Artetxe et~al.(2018{\natexlab{b}})Artetxe, Labaka, Agirre, and
  Cho}]{artetxe2018unmt}
Mikel Artetxe, Gorka Labaka, Eneko Agirre, and Kyunghyun Cho.
  2018{\natexlab{b}}.
\newblock \href {https://openreview.net/pdf?id=Sy2ogebAW} {Unsupervised neural
  machine translation}.
\newblock In \emph{Proceedings of the 6th International Conference on Learning
  Representations (ICLR 2018)}.

\bibitem[{Barrault et~al.(2019)Barrault, Bojar, Costa-juss{\`a}, Federmann,
  Fishel, Graham, Haddow, Huck, Koehn, Malmasi, Monz, M{\"u}ller, Pal, Post,
  and Zampieri}]{barrault2019findings}
Lo{\"\i}c Barrault, Ond{\v{r}}ej Bojar, Marta~R. Costa-juss{\`a}, Christian
  Federmann, Mark Fishel, Yvette Graham, Barry Haddow, Matthias Huck, Philipp
  Koehn, Shervin Malmasi, Christof Monz, Mathias M{\"u}ller, Santanu Pal, Matt
  Post, and Marcos Zampieri. 2019.
\newblock \href {https://doi.org/10.18653/v1/W19-5301} {Findings of the 2019
  conference on machine translation ({WMT}19)}.
\newblock In \emph{Proceedings of the Fourth Conference on Machine Translation
  (Volume 2: Shared Task Papers, Day 1)}, pages 1--61, Florence, Italy.
  Association for Computational Linguistics.

\bibitem[{Burlot and Yvon(2018)}]{burlot2018using}
Franck Burlot and Fran{\c{c}}ois Yvon. 2018.
\newblock \href {https://doi.org/10.18653/v1/W18-6315} {Using monolingual data
  in neural machine translation: a systematic study}.
\newblock In \emph{Proceedings of the Third Conference on Machine Translation:
  Research Papers}, pages 144--155, Brussels, Belgium. Association for
  Computational Linguistics.

\bibitem[{Conneau and Lample(2019)}]{conneau2019crosslingual}
Alexis Conneau and Guillaume Lample. 2019.
\newblock \href
  {http://papers.nips.cc/paper/8928-cross-lingual-language-model-pretraining.pdf}
  {Cross-lingual language model pretraining}.
\newblock In \emph{Advances in Neural Information Processing Systems 32}, pages
  7059--7069.

\bibitem[{Dyer et~al.(2013)Dyer, Chahuneau, and Smith}]{dyer2013simple}
Chris Dyer, Victor Chahuneau, and Noah~A. Smith. 2013.
\newblock \href {https://www.aclweb.org/anthology/N13-1073} {A simple, fast,
  and effective reparameterization of {IBM} model 2}.
\newblock In \emph{Proceedings of the 2013 Conference of the North {A}merican
  Chapter of the Association for Computational Linguistics: Human Language
  Technologies}, pages 644--648, Atlanta, Georgia. Association for
  Computational Linguistics.

\bibitem[{Edunov et~al.(2018)Edunov, Ott, Auli, and
  Grangier}]{edunov2018understanding}
Sergey Edunov, Myle Ott, Michael Auli, and David Grangier. 2018.
\newblock \href {https://doi.org/10.18653/v1/D18-1045} {Understanding
  back-translation at scale}.
\newblock In \emph{Proceedings of the 2018 Conference on Empirical Methods in
  Natural Language Processing}, pages 489--500, Brussels, Belgium. Association
  for Computational Linguistics.

\bibitem[{Fadaee and Monz(2018)}]{fadaee2018backtranslation}
Marzieh Fadaee and Christof Monz. 2018.
\newblock \href {https://doi.org/10.18653/v1/D18-1040} {Back-translation
  sampling by targeting difficult words in neural machine translation}.
\newblock In \emph{Proceedings of the 2018 Conference on Empirical Methods in
  Natural Language Processing}, pages 436--446, Brussels, Belgium. Association
  for Computational Linguistics.

\bibitem[{He et~al.(2016)He, Xia, Qin, Wang, Yu, Liu, and Ma}]{he2016dual}
Di~He, Yingce Xia, Tao Qin, Liwei Wang, Nenghai Yu, Tie-Yan Liu, and Wei-Ying
  Ma. 2016.
\newblock \href
  {http://papers.nips.cc/paper/6469-dual-learning-for-machine-translation.pdf}
  {Dual learning for machine translation}.
\newblock In \emph{Advances in Neural Information Processing Systems 29}, pages
  820--828.

\bibitem[{Heafield et~al.(2013)Heafield, Pouzyrevsky, Clark, and
  Koehn}]{heafield2013scalable}
Kenneth Heafield, Ivan Pouzyrevsky, Jonathan~H. Clark, and Philipp Koehn. 2013.
\newblock \href {https://www.aclweb.org/anthology/P13-2121} {Scalable modified
  {K}neser-{N}ey language model estimation}.
\newblock In \emph{Proceedings of the 51st Annual Meeting of the Association
  for Computational Linguistics (Volume 2: Short Papers)}, pages 690--696,
  Sofia, Bulgaria. Association for Computational Linguistics.

\bibitem[{Hoang et~al.(2018)Hoang, Koehn, Haffari, and
  Cohn}]{hoang2018iterative}
Vu~Cong~Duy Hoang, Philipp Koehn, Gholamreza Haffari, and Trevor Cohn. 2018.
\newblock \href {https://doi.org/10.18653/v1/W18-2703} {Iterative
  back-translation for neural machine translation}.
\newblock In \emph{Proceedings of the 2nd Workshop on Neural Machine
  Translation and Generation}, pages 18--24, Melbourne, Australia. Association
  for Computational Linguistics.

\bibitem[{Koehn et~al.(2007)Koehn, Hoang, Birch, Callison-Burch, Federico,
  Bertoldi, Cowan, Shen, Moran, Zens, Dyer, Bojar, Constantin, and
  Herbst}]{koehn2007moses}
Philipp Koehn, Hieu Hoang, Alexandra Birch, Chris Callison-Burch, Marcello
  Federico, Nicola Bertoldi, Brooke Cowan, Wade Shen, Christine Moran, Richard
  Zens, Chris Dyer, Ond{\v{r}}ej Bojar, Alexandra Constantin, and Evan Herbst.
  2007.
\newblock \href {https://www.aclweb.org/anthology/P07-2045} {{M}oses: Open
  source toolkit for statistical machine translation}.
\newblock In \emph{Proceedings of the 45th Annual Meeting of the Association
  for Computational Linguistics Companion Volume Proceedings of the Demo and
  Poster Sessions}, pages 177--180, Prague, Czech Republic. Association for
  Computational Linguistics.

\bibitem[{Koehn et~al.(2003)Koehn, Och, and Marcu}]{koehn2003statistical}
Philipp Koehn, Franz~J. Och, and Daniel Marcu. 2003.
\newblock \href {https://www.aclweb.org/anthology/N03-1017} {Statistical
  phrase-based translation}.
\newblock In \emph{Proceedings of the 2003 Human Language Technology Conference
  of the North {A}merican Chapter of the Association for Computational
  Linguistics}, pages 127--133.

\bibitem[{Lample et~al.(2018{\natexlab{a}})Lample, Conneau, Denoyer, and
  Ranzato}]{lample2018unsupervised}
Guillaume Lample, Alexis Conneau, Ludovic Denoyer, and Marc'Aurelio Ranzato.
  2018{\natexlab{a}}.
\newblock \href {https://openreview.net/pdf?id=rkYTTf-AZ} {Unsupervised machine
  translation using monolingual corpora only}.
\newblock In \emph{Proceedings of the 6th International Conference on Learning
  Representations (ICLR 2018)}.

\bibitem[{Lample et~al.(2018{\natexlab{b}})Lample, Ott, Conneau, Denoyer, and
  Ranzato}]{lample2018phrase}
Guillaume Lample, Myle Ott, Alexis Conneau, Ludovic Denoyer, and Marc{'}Aurelio
  Ranzato. 2018{\natexlab{b}}.
\newblock \href {https://doi.org/10.18653/v1/D18-1549} {Phrase-based {\&}
  neural unsupervised machine translation}.
\newblock In \emph{Proceedings of the 2018 Conference on Empirical Methods in
  Natural Language Processing}, pages 5039--5049, Brussels, Belgium.
  Association for Computational Linguistics.

\bibitem[{Liu et~al.(2020)Liu, Gu, Goyal, Li, Edunov, Ghazvininejad, Lewis, and
  Zettlemoyer}]{liu2020multilingual}
Yinhan Liu, Jiatao Gu, Naman Goyal, Xian Li, Sergey Edunov, Marjan
  Ghazvininejad, Mike Lewis, and Luke Zettlemoyer. 2020.
\newblock Multilingual denoising pre-training for neural machine translation.
\newblock \emph{arXiv preprint arXiv:2001.08210}.

\bibitem[{Marie and Fujita(2018)}]{marie2018unsupervised}
Benjamin Marie and Atsushi Fujita. 2018.
\newblock Unsupervised neural machine translation initialized by unsupervised
  statistical machine translation.
\newblock \emph{arXiv preprint arXiv:1810.12703}.

\bibitem[{Och(2003)}]{och2003MERT}
Franz~Josef Och. 2003.
\newblock \href {https://doi.org/10.3115/1075096.1075117} {Minimum error rate
  training in statistical machine translation}.
\newblock In \emph{Proceedings of the 41st Annual Meeting of the Association
  for Computational Linguistics}, pages 160--167, Sapporo, Japan. Association
  for Computational Linguistics.

\bibitem[{Ott et~al.(2018)Ott, Edunov, Grangier, and Auli}]{ott2018scaling}
Myle Ott, Sergey Edunov, David Grangier, and Michael Auli. 2018.
\newblock \href {https://doi.org/10.18653/v1/W18-6301} {Scaling neural machine
  translation}.
\newblock In \emph{Proceedings of the Third Conference on Machine Translation:
  Research Papers}, pages 1--9, Brussels, Belgium. Association for
  Computational Linguistics.

\bibitem[{Poncelas et~al.(2018)Poncelas, Shterionov, Way, Wenniger, and
  Passban}]{poncelas2018investigating}
Alberto Poncelas, Dimitar Shterionov, Andy Way, Gideon Maillette de~Buy
  Wenniger, and Peyman Passban. 2018.
\newblock Investigating backtranslation in neural machine translation.
\newblock \emph{arXiv preprint arXiv:1804.06189}.

\bibitem[{Sennrich et~al.(2016)Sennrich, Haddow, and
  Birch}]{sennrich2016improving}
Rico Sennrich, Barry Haddow, and Alexandra Birch. 2016.
\newblock \href {https://doi.org/10.18653/v1/P16-1009} {Improving neural
  machine translation models with monolingual data}.
\newblock In \emph{Proceedings of the 54th Annual Meeting of the Association
  for Computational Linguistics (Volume 1: Long Papers)}, pages 86--96, Berlin,
  Germany. Association for Computational Linguistics.

\bibitem[{Song et~al.(2019)Song, Tan, Qin, Lu, and Liu}]{song2019mass}
Kaitao Song, Xu~Tan, Tao Qin, Jianfeng Lu, and Tie-Yan Liu. 2019.
\newblock \href {http://proceedings.mlr.press/v97/song19d.html} {{MASS}: Masked
  sequence to sequence pre-training for language generation}.
\newblock In \emph{Proceedings of the 36th International Conference on Machine
  Learning}, volume~97 of \emph{Proceedings of Machine Learning Research},
  pages 5926--5936, Long Beach, California, USA. PMLR.

\end{thebibliography}
\bibliographystyle{acl_natbib}

\end{document}